\documentclass[conference]{IEEEtran}
\IEEEoverridecommandlockouts
\usepackage{cite}
\usepackage{amsmath,amssymb,amsfonts}
\usepackage{graphicx}
\usepackage{textcomp}
\usepackage{xcolor}
\usepackage{wrapfig}

\usepackage{graphicx}
\usepackage{comment}
\usepackage{amsmath,amssymb} 
\usepackage{color}

\usepackage{url}
\usepackage{booktabs}
\usepackage{nicefrac}
\usepackage{microtype}
\usepackage{epsfig}
\usepackage[tight,footnotesize]{subfigure}
\usepackage{amsmath}
\usepackage{amssymb}
\usepackage{amsthm}
\usepackage{bm}
\usepackage{multirow}
\usepackage{makecell}
\usepackage{footmisc}
\usepackage{marvosym}
\usepackage{latexsym}
\usepackage[utf8]{inputenc}
\usepackage{kotex}
\usepackage{titlesec}
\usepackage{caption}
\usepackage{microtype}
\usepackage{graphicx}
\usepackage{subfigure}
\usepackage{booktabs} 
\usepackage{amsfonts}
\usepackage{breqn}
\usepackage{algorithm,algpseudocode}

\newcommand{\ie}{\textit{i}.\textit{e}., }

\def\BibTeX{{\rm B\kern-.05em{\sc i\kern-.025em b}\kern-.08em
    T\kern-.1667em\lower.7ex\hbox{E}\kern-.125emX}}
    
\begin{document}

\title{Examining the Robustness of Spiking Neural Networks on Non-ideal Memristive Crossbars
}
\author{
\IEEEauthorblockN{  Abhiroop Bhattacharjee$^{*}$\thanks{\hspace{-3mm}$^*$ These authors have contributed equally to this work.}, Youngeun Kim$^{*}$, Abhishek Moitra, and  Priyadarshini Panda} 
\IEEEauthorblockA{
Department of Electrical Engineering, Yale University, USA\\
\{abhiroop.bhattacharjee, youngeun.kim, abhishek.moitra,  priya.panda\}@yale.edu}
}



\maketitle

\begin{abstract} 
Spiking Neural Networks (SNNs) have recently emerged as the low-power alternative to Artificial Neural Networks (ANNs) owing to their asynchronous, sparse, and binary information processing.
To improve the energy-efficiency and throughput, SNNs can be implemented on memristive crossbars where Multiply-and-Accumulate (MAC) operations are realized in the analog domain using emerging Non-Volatile-Memory (NVM) devices.
Despite the compatibility of SNNs with memristive crossbars, there is little attention to study on the effect of intrinsic crossbar non-idealities and stochasticity on the performance of SNNs.
In this paper, we conduct a comprehensive analysis of the robustness of SNNs on non-ideal crossbars.
We examine SNNs trained via learning algorithms such as, surrogate gradient and ANN-SNN conversion. 
Our results show that repetitive crossbar computations across multiple time-steps induce error accumulation, resulting in a huge performance drop during SNN inference.
We further show that SNNs trained with a smaller number of time-steps achieve better accuracy when deployed on memristive crossbars. 
\end{abstract}


\section{Introduction}
The  previous  decade  has  witnessed  the  rise  of  Spiking Neural Networks  (SNNs) in the context of low-power machine intelligence \cite{roy2019towards,christensen20222022}. SNNs, unlike Artificial Neural Networks (ANNs), process visual information with discrete binary spikes or events over multiple time-steps resulting in asynchronous event-driven processing. Recent works have shown that the event-driven behaviour of SNNs can be efficiently implemented on emerging neuromorphic hardware to yield 1-2 orders of magnitude of energy-efficiency compared to that of ANNs on static image classification problems \cite{davies2018loihi,furber2014spinnaker,akopyan2015truenorth}. To this end, memristive crossbars, built using Non-Volatile-Memory (NVM) devices, have emerged as a fast, compact and energy-efficient solution to implementing neural networks for In-Memory Computing (IMC) in the analog domain \cite{chakraborty2020pathways, ankit2017resparc}.

\begin{figure}[t]
    \centering
    \includegraphics[width=\linewidth]{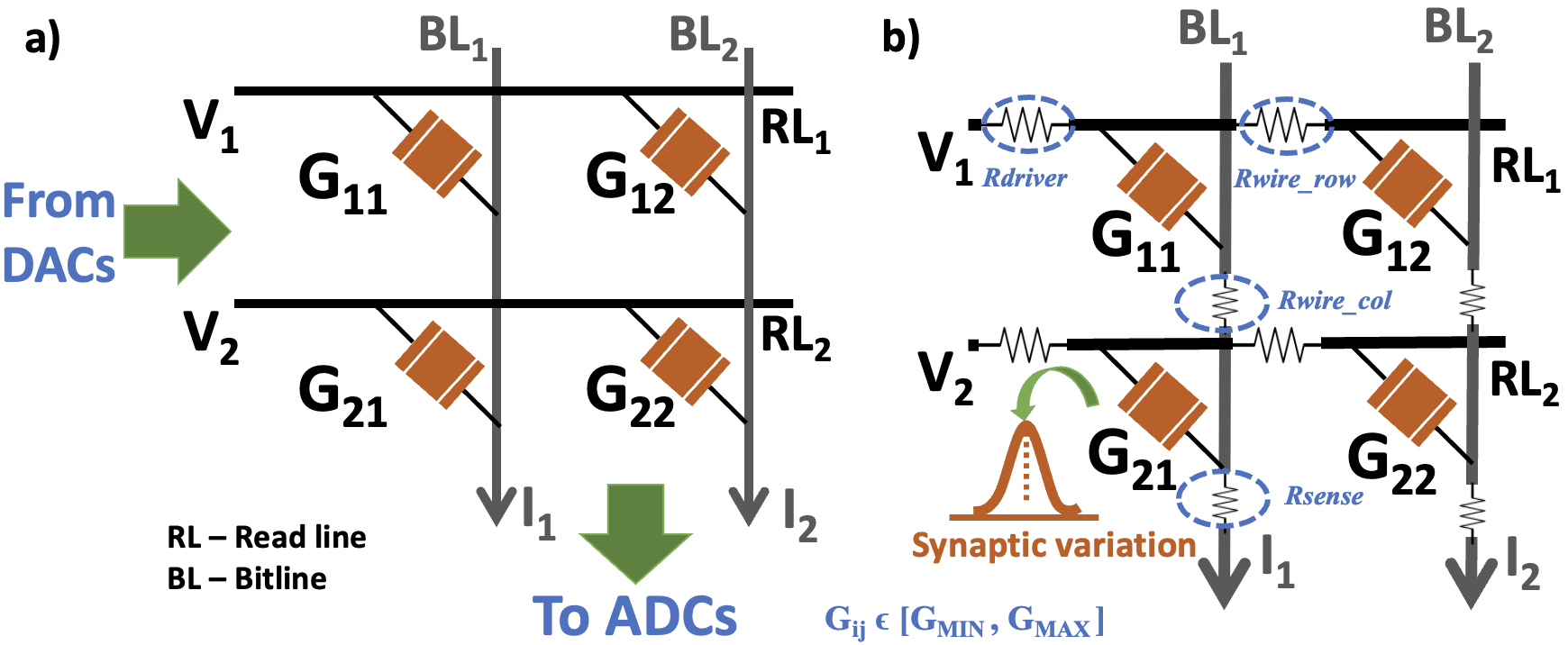}
    \caption{(a) Ideal crossbar with input voltages $V_{i}$, synaptic conductances $G_{ij}$ and output currents $I_{j} = \sum_{i}^{}{G_{ij} V_i}$; (b) Non-ideal crossbar with the interconnect and synaptic non-idealities annotated.}
    \label{xbar}
\end{figure}

However, there are several pitfalls to IMC using analog crossbars. Crossbars possess several non-idealities such as, interconnect parasitics, non-linearities/variations in the synapses, etc. \cite{jain2020rxnn, chakraborty2020geniex} that result in imprecise dot-product currents, leading to performance (accuracy) degradation upon mapping neural networks.
Many works have modelled these non-idealities to study their impact on the performance and robustness of crossbar-mapped ANNs \cite{bhattacharjee2021neat, liu2015vortex, bhattacharjee2021efficiency, bhattacharjee2020rethinking,bhattacharjee2021switchx}. Although SNNs, as compared to ANNs, have been shown to be more robust to input noise  owing to inherent stochasticity in the spike-encoding methods \cite{sharmin2020inherent}, there is no work that assesses the impact of crossbar non-idealities on the performance (robustness) of SNN models. 
To this end, our work is aimed at understanding- \textit{How robust is an SNN, evaluated over multiple time-steps, with respect to its ANN counterpart when mapped onto non-ideal IMC crossbars?} 
Our findings indicate that, for an SNN trained with gradient backpropagation, repetitive crossbar operations across multiple time-steps induce \textit{error accumulation} due to non-idealities that has detrimental impact on the performance of SNNs on crossbars as compared to ANNs.
On the other hand, an SNN converted from pre-trained ANN is not affected by  the number of time-steps since the converted SNN is the approximated version of the ANN model.


In summary, the key contributions of this work are as follows:
\begin{itemize}
    \item This work for the first time compares the robustness of SNNs on non-ideal memristive crossbars against corresponding ANN models. We examine  SNNs  trained  from various  learning  algorithms  such  as, \textit{Surrogate  Gradient} (SG) learning  \cite{wu2018spatio}, \textit{ANN-SNN  conversion} \cite{sengupta2019going}, and \textit{Batch Normalization Through Time} (BNTT) \cite{bntt}. 
    
    \item We conduct experiments with VGG5 and VGG9 network architectures for ANNs and SNNs using benchmark datasets, namely MNIST and CIFAR10, to show that SNNs underperform with respect to ANNs when mapped on non-ideal crossbars of size 64$\times$64 and 32$\times$32.
    
    \item We also perform ablation studies using SG-trained VGG9/CIFAR10 SNNs with different time-steps (10, 20 and 30).
    We find that a reduction in the number of time-steps improves the performance of SNNs on non-ideal crossbars. 
    
    \item We introduce noise-aware batch normalization adaptation technique
    to mitigate the vulnerability of BNTT-trained SNNs on non-ideal crossbars.
\end{itemize}

\section{Related Works}

There has been a plethora of works that have proposed hardware accelerator designs to carry out SNN inference showing a high degree of parallelism, throughput, and energy-efficiency \cite{roy2019towards, panda2020toward, ankit2017resparc, davies2018loihi, balaji2019mapping, akopyan2015truenorth}. These include inference accelerators with a fully-digital architecture, such as IBM’s TrueNorth processor \cite{akopyan2015truenorth}, as well one in which synaptic computational cores comprise of analog memristive crossbars, such as RESPARC \cite{ankit2017resparc}. A recent work \cite{balaji2019mapping} proposed a methodology to map spiking neurons on analog crossbar architectures such that neurons having higher spike activities are placed close together to reduce communication energy overheads and better energy-efficiency. In summary, the primary focus of majority of these works has been to facilitate sparse event-driven spike communications with the key objective of improving the energy-efficiency of the deployed SNNs. But, none of the works on crossbar-based accelerators for SNNs account for the impact of the inexorable parasitic non-idealities that can detrimentally impact the performance of the mapped SNN architectures. Unless it is ascertained that the trained SNN models can yield a descent performance on mapping onto non-ideal crossbar arrays, the energy-efficiency advantages extracted from SNNs become inconsequential. 

There have been many prior works that have included non-idealities during the inference of ANNs on crossbars \cite{bhattacharjee2021efficiency, bhattacharjee2021neat, bhattacharjee2022examining, chakraborty2020geniex, liu2015vortex}. These include \textit{Crosssim} \cite{plimpton2016crosssim} and \textit{Neurosim} \cite{chen2018neurosim} platforms that include device-to-device variations in the NVM synapses during inference. However, the resistive crossbar non-idealities (interconnect parasitics) that lead to IR-drops during the dot-product operations are not modelled by these platforms. To this end, a recent work called \textit{RxNN} \cite{jain2020rxnn} splits and maps an ANNs computations into crossbar operations and proposes a fast crossbar model (FCM) to accurately capture the impact of resistive crossbar non-idealities during inference. The non-ideality integration framework employed in this work for evaluating SNN models is similar to the \textit{RxNN} method for ANNs, and can model the impact of both resistive and device-to-device non-idealities. With this framework, we conduct extensive experiments to show the community where the performance of SNNs on analog crossbars stands in comparison to their ANN counterparts.


\section{Background}

\subsection{Memristive Crossbars and Non-Idealities}
\begin{figure*}[t]
    \centering
    \includegraphics[width=1.0\linewidth]{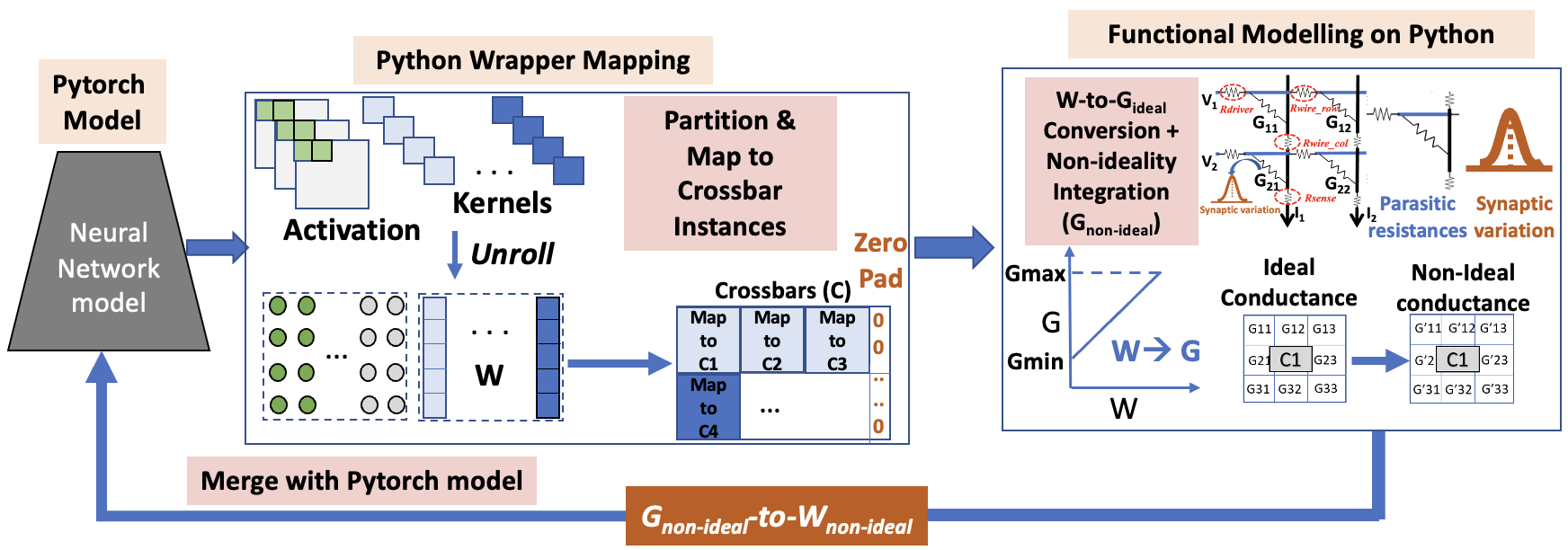}
    \caption{Hardware evaluation framework in Python to map ANNs or SNNs on non-ideal crossbars, followed by inference on the crossbar-mapped models. It includes partitioning the original weight matrices into numerous crossbars, followed by conversion of the weights to conductances $G_{(ideal)}$. Thereafter, $G_{(ideal)}$ is transformed to $G_{(non-ideal)}$ using linear algebra and circuit laws and finally, we extract the non-ideal weights for inference.}
    \label{hw_framework}
\end{figure*}
Memristive crossbars are used to realize Multiply-and-Accumulate (MAC) operations on IMC hardware in an analog manner. Crossbars receive the input activations of a neural network as analog voltages from Digital-to-Analog converters (DACs) and produce currents analogous to the outputs of MAC operations that are sensed by Analog-to-Digital converters (ADCs). Ideally, the MAC operations occur using Ohm's Law and Kirchoff's current law with the interaction of the input voltages and the memristive conductances for the synapses (programmed between $G_{MIN}$ and $G_{MAX}$) as shown in Fig. \ref{xbar}(a). However, the analog nature of the computation leads to various non-idealities, such as, circuit-level interconnect parasitics and synaptic-level non-linearities or variations~\cite{bhattacharjee2021neat, chakraborty2020geniex, jain2020rxnn}. 

    

Fig. \ref{xbar}(b) describes the equivalent circuit for a memristive crossbar consisting of various circuit-level and device-level non-idealities, \textit{viz.} $Rdriver$, $Rwire\_row$, $Rwire\_col$ and $Rsense$ (interconnect parasitics), modelled as parasitic resistances and variations/non-linearities in the memristive synapses. The impact of these non-idealities or hardware noise can be incorporated by transforming the ideal memristive conductances $G_{ij(ideal)}$ (obtained from the weights of a neural network) to non-ideal conductances $G_{ij(non-ideal)}$ using linear algebraic operations and circuit laws. Consequently, the net output current sensed at the end of the crossbar-columns ($I_{non-ideal}$) deviates from its ideal value ($I_{ideal}$). 

\subsection{Spiking Neural Networks}

SNNs \cite{roy2019towards,diehl2015unsupervised} have gained attention due to their potential energy-efficiency compared to standard ANNs. 
The main feature of SNNs is the type of neural activation function for temporal signal processing, which is different from a ReLU activation for ANNs.
A Leak-Integrate-and-Fire (LIF) neuron   is commonly used as an activation function for SNNs.
The LIF neuron $i$ has a membrane potential $u_{i}^{t}$ which accumulates the asynchronous spike inputs, which can be formulated as follows:
\begin{equation}
    u_i^t = \lambda  u_i^{t-1} + \sum_j w_{ij}o^t_j.
    \label{eq:LIF}
\end{equation}
Here, $t$ stands for time-step, and $w_{ij}$ is for weight connections between neuron $i$ and neuron $j$. Also, $\lambda$ is a leak factor.
The  LIF neuron $i$ accumulates membrane potential and generates a spike output $o_i^{t}$ whenever membrane potential exceeds the threshold $\theta$:
\begin{equation}
    o^{t}_i =
\begin{cases}
 1,          & \text{if $u_i^{t} >\theta$},  \\
    0
    & \text{otherwise.} 
\end{cases}
\label{eq:firing}
\end{equation}
The membrane potential is reset to zero after firing.
This integrate-and-fire behavior of an LIF neuron generates a non-differentiable function, which is difficult to be used with standard backpropagation.

To address this, various training algorithms for SNNs have been studied in the past few decades.
ANN-SNN conversion methods \cite{sengupta2019going,diehl2015fast,han2020deep,li2021free,rueckauer2017conversion} convert pre-trained ANNs to SNNs using weight (or threshold) scaling   in order to approximate ReLU activation with LIF activation.
They can leverage well-established ANN training methods, resulting in high accuracy on complex datasets.
On the other hand, surrogate gradient learning addresses the non-differentiability problem of a LIF neuron by approximating the backward gradient function \cite{wu2018spatio}.
Surrogate gradient learning can directly learn from the spikes, in a smaller number of time-steps. 
Based on the surrogate learning, several advanced algorithms have been proposed. 
Tandem learning \cite{wu2019tandem,wu2020progressive}
designs an auxiliary ANN in order to stabilise error back-propagation for SNN training.
Moreover, a line of work \cite{fang2021incorporating,rathi2021diet} directly train membrane decay (or firing threshold) in an LIF neuron with standard weight update, resulting in a better representation power of SNNs. 
Batch Normalization (BN)  technique  \cite{ioffe2015batch}  has been applied to accelerate the training process of SNNs \cite{ledinauskas2020training,bntt,zheng2020going}. 
In this work, we measure the impact of crossbar non-idealities during the inference of SNNs trained via both ANN-SNN conversion and surrogate gradients.


\section{Robustness of SNNs on Memristive Crossbars}

\begin{algorithm}[t]\small
        \caption{ANN-SNN Conversion \cite{sengupta2019going}}
      \textbf{Input}: data set ($X$); label set ($Y$); max time-step ($T$); network depth ($L$); pre-trained ANN model ($ANN$); SNN model ($SNN$)   \\
      \textbf{Output}:  converted SNN network
      \begin{algorithmic}[1]
        %
        \State {$SNN.weights \leftarrow ANN.weights$}\Comment{Copy  weights}
        \State {$SNN.th$ $\leftarrow$ 0}\Comment{Initialize firing threshold}
        \For{$l \gets 1$ to $L-1$}
            \For{$t \gets 1$ to $T$}  
                \State{$O^{t}$ $\leftarrow$ PoissonGenerator(X)}
                \For{$l' \gets 1$ to $l$} 
                    \If {$l' < l$}
                        \State{$(O^t_{l}, U_l^{t}) \leftarrow 
                        ( U_l^{t-1}, W_{l}, O^{t}_{l-1})$}\Comment{Forward (Eq. \ref{eq:LIF})}
                    \Else
                        \State{$SNN_{l}.th$ $\leftarrow$ max($SNN_{l}.th$, ${W_{l}O^{t}_{l-1}}$)}
                    \EndIf
                \EndFor
            \EndFor
        \EndFor
      \end{algorithmic}
          \label{algorithm: overall}

\end{algorithm}

\begin{algorithm}[t]\small
    \caption{Training process with Surrogate Gradient}
   \textbf{Input}: data set ($X$); label set ($Y$); max time-step ($T$) \\
  \textbf{Output}:  updated network weights 
  \begin{algorithmic}[1]
    %
    \For{$i \gets 1$ to $max\_iter$}
        \State {fetch a mini batch X}
        \For{$t \gets 1$ to $T$}  
            \State{O $\leftarrow$ PoissonGenerator(X)}
            \For{$l \gets 1$ to $L-1$} 
                \State{$(O^t_{l}, U_l^{t}) \leftarrow (\lambda, U_l^{t-1}, (W_{l}, O^{t}_{l-1}))$}
            \EndFor
            \State{\% For the final layer $L$, accumulate the voltage}
            \State{$U_L^{t} \hspace{-1mm} \leftarrow  \hspace{-1mm} ( U_L^{t-1},  (W_{l}, O^{t}_{L-1}))$ }
        \EndFor
        \State{\% Calculate the loss and back-propagation}
        \State{$L \leftarrow (U_L^T, Y)$}
    \EndFor
  \end{algorithmic}
      \label{algorithm: overall}
\end{algorithm}

\subsection{Optimizing SNNs}
\label{ssec:SNNopt-algorithm}

Here, we first describe two representative SNN training algorithms in detail used in our experiments.\\

\noindent\textbf{{ANN-SNN conversion}}:
We use an iterative layer-wise ANN-SNN conversion method proposed in \cite{sengupta2019going}.
This method normalizes the firing thresholds in SNNs to approximate float activation value in a pre-trained ANN model.
The overall conversion algorithm is presented in Algorithm 1.
Firstly, we initialize SNNs using the weights from a pre-trained ANN (line 1).
After that, we search the maximum activation value across all time-steps in a layer, and set the firing threshold with searched activation value (line 3-14). 
The conversion process starts from the first layer and sequentially goes through deeper layers. \\

\begin{table*}[t!]
    \addtolength{\tabcolsep}{2.5pt}
    \centering
    \caption{Classification accuracy (\%) of ANNs and SNNs on MNIST and CIFAR10 datasets. \textit{SW accuracy} shows the clean performance achieved on a GPU machine.   \textit{HW accuracy} takes into account   non-idealities in $64 \times 64$ crossbars during inference. The symbol $\Delta$ represents a relative accuracy (performance) drop, \ie $\frac{SW accuracy - HW accuracy}{SW accuracy} \times 100$, that denotes our metric to measure ANN and SNN performances.}
    \label{table:accuracy_cifar100}
    \resizebox{\linewidth}{!}{%
    \begin{tabular}{lcccccc}
        \toprule
        Method &\textrm{Architecture} 
        & \textrm{Dataset} &  \textrm{Time-steps} &
        \textrm{SW accuracy (\%)} &
        \textrm{HW accuracy (\%)} &\textrm{$\Delta$ (\%)} \\
         \midrule
     ANN &  VGG5 & MNIST & - & 99.34 & 99.20 & 0.14 \\
     ANN &  VGG9 & CIFAR10 & - & 91.90 & 86.29 & 6.10 \\
     Conversion \cite{sengupta2019going} &  VGG5 & MNIST & 50 & 99.26 & 99.07 & 0.19 \\
     Conversion \cite{sengupta2019going} &  VGG9 & CIFAR10 & 500 & 91.02 & 86.40 & 5.07 \\
     Surrogate Gradient \cite{wu2018spatio} &  VGG5 & MNIST & 25 & 99.32 & 98.52 & 0.80 \\
     Surrogate Gradient \cite{wu2018spatio} &  VGG9 & CIFAR10 & 30 & 86.70 & 14.94 & 82.76 \\
     BNTT \cite{bntt} &  VGG5 & MNIST & 10 & 99.45 & 11.35 & 88.58 \\
     BNTT \cite{bntt} &  VGG9 & CIFAR10 & 20 & 90.43 & 10.01 & 88.93 \\
        \bottomrule
        \\
    \end{tabular}%
    }
    \label{table:performance}
\end{table*}

\noindent\textbf{{Surrogate Gradient Backpropagation}}:
Surrogate gradient learning approximates the non-differentiable firing behavior of LIF neurons by using surrogate function.
Let $o^t_i$ and $u^t_i$ be output spikes and membrane potential at time-step $t$ of neuron $i$, respectively.
To calculate gradients, we use piece-wise linear backward function: 
\begin{equation}
    \frac{\partial o_i^t}{\partial u_i^t} = \max \{0, 1-  \ | \frac{u_i^t - \theta}{\theta} \ | \},
    \label{eq:approx_grad_function}
\end{equation}
where, $\theta$ represents the firing threshold.
With such an approximated function, surrogate gradient based backpropagation learning can be implemented on machine learning frameworks like PyTorch \cite{paszke2017automatic}.
In our experiments, we use spatio-temporal back-propagation (STBP) \cite{wu2018spatio}.
They that accumulates the gradients over spatial and temporal dimensions which can be formulated as follows:
\begin{equation}
      \frac{\partial L}{\partial W_l} =
\begin{cases}
 \sum_{t}(\frac{\partial L}{\partial O_l^t}\frac{\partial O_l^t}{\partial U_l^t} + \frac{\partial L}{\partial U_l^{t+1}}  \frac{\partial U_l^{t+1}}{\partial U_l^{t}})
 \frac{\partial U_l^t}{\partial W_l},  & \text{if $l:$ hidden } \\
    \frac{\partial L}{\partial U_l^T}\frac{\partial U_l^T}{\partial W_l}.
    & \text{if $l:$ output} 
\end{cases}
\label{eq:delta_W}
\end{equation}
Here, $O^t_l$ and $U^t_l$ are output spikes and membrane potential at time-step $t$ for layer $l$, respectively.
The details are presented in Algorithm 2.
%

\subsection{Experimental settings}

\noindent\textbf{Hardware:} 
In this work, we use a crossbar-based hardware evaluation framework (see Fig. \ref{hw_framework}) \cite{bhattacharjee2020rethinking} to include the impact of crossbar non-idealities to the weights of a trained ANN or SNN during inference. The entire framework is written in Python. It involves a Python wrapper that reshapes the 4D convolutional weight-matrices of each layer of trained ANNs or SNNs into 2D matrices ($W$) consisting of ideal weights. Thereafter, the matrices are partitioned into multiple crossbar instances (of a given size), followed by conversion of the ideal weights into synaptic conductances. Next, we model the resistive crossbar non-idealities using circuit laws and linear algebraic operations in Python \cite{jain2020rxnn} and obtain the non-ideal conductance matrices. Finally, the non-ideal synaptic conductances are transformed into non-ideal weights which are then integrated into the original Pytorch based ANN or SNN model to conduct inference. Note, we follow an \textit{NVM device agnostic} approach for analyzing the impact of intrinsic circuit-level and synaptic crossbar non-idealities on the performance of SNNs or ANNs during inference. The ON/OFF ratio for the NVM devices in the crossbars is considered to be 10 (\textit{i.e.,} $R_{MIN} = 20 k\Omega$ and $R_{MAX} = 200 k\Omega$), typical for ReRAM devices. The resistive non-idealities (see Fig. \ref{xbar}(b)) are as follows: $Rdriver = 1 k\Omega$, $Rwire\_row = 5 \Omega$, $Rwire\_col = 10 \Omega$ and $Rsense = 1 k\Omega$. The synaptic variations are modelled as a Gaussian variation in the synaptic conductances with $\sigma/\mu = 10\%$ \cite{bhattacharjee2021neat}.\\

\noindent\textbf{Software:} 
 In our experiments, we use two convolutional architectures (\ie VGG5 and VGG9) on two public datasets (\ie MNIST and CIFAR10).
{MNIST} \cite{lecun1998gradient} contains gray-scale images of size 28 $\times$ 28. 
{CIFAR10} \cite{krizhevsky2009learning} consists of 60,000 RGB color images of size 32 $\times$ 32 (50,000 for training / 10,000 for testing) with 10 categories. 
Our implementation is carried out on the PyTorch framework \cite{paszke2017automatic}. 
Total number of training epochs are set to 60 and 100 for MNIST and CIFAR10 datasets, respectively.
During training, we utilize step-wise learning rate scheduling with a decay factor of 10 at 50\% and 75\% of the total epochs. 
We train the networks with Adam optimizer with an initial learning rate $1e-4$.

\begin{figure}[t]
\begin{center}
\def\arraystretch{0.5}
\begin{tabular}{@{}c@{\hskip 0.000\linewidth}c@{}c}
\includegraphics[width=0.5\linewidth]{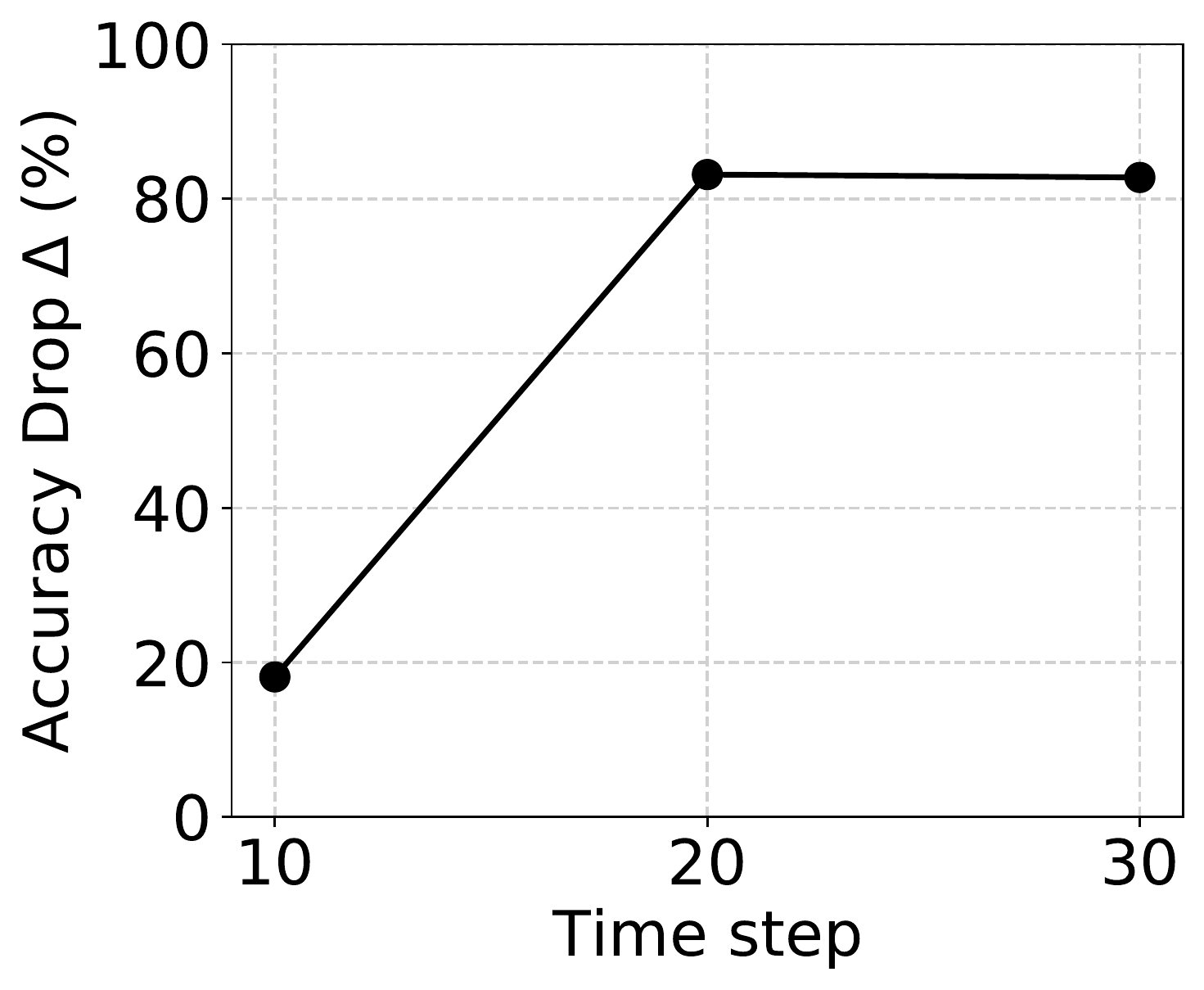} &
\includegraphics[width=0.5\linewidth]{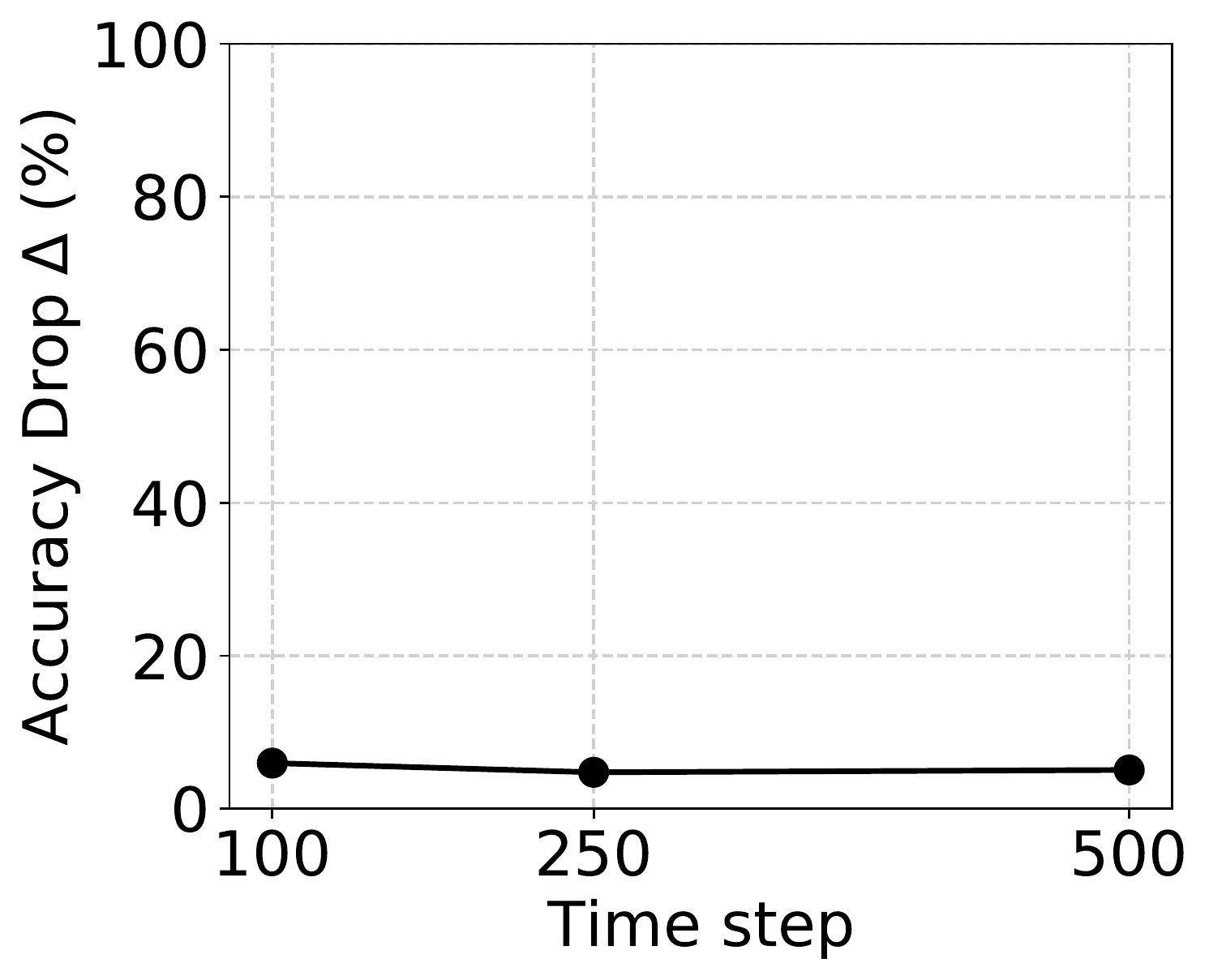} 
\\
{\hspace{7mm} (a) Surrogate Gradients } & {\hspace{4mm} (b) ANN-SNN Conversion }\\
\end{tabular}
\end{center}
\caption{Accuracy drop ($\Delta$) for VGG9/CIFAR10 SNNs with respect to time-steps.
}
\vspace{-4mm}
\label{fig:timesteps_abl}
\end{figure}


\begin{table*}[t!]
    \addtolength{\tabcolsep}{2.5pt}
    \centering
    \caption{Table with Classification accuracy (\%) of ANNs and SNNs with VGG9 architecture on CIFAR10 dataset showing \textit{SW accuracy}, \textit{HW accuracy} and $\Delta$ for 32$\times$32 and  64$\times$64 non-ideal crossbars during inference.}
    \label{table:accuracy_cifar100}
    \resizebox{0.8\linewidth}{!}{%
    \begin{tabular}{lccccc}
        \toprule
        Method &\textrm{Time-steps} & \textrm{Crossbar size} &
        \textrm{SW acc. (\%)} &
        \textrm{HW acc. (\%)} &\textrm{$\Delta$ (\%)} \\
         \midrule
     ANN  & - & 32$\times$32  & 91.90 & 90.46 & 1.56 \\
          ANN  & - & 64$\times$64  & 91.90 & 86.29  & 6.10 \\
     Surrogate \cite{wu2018spatio} & 20& 32$\times$32  & 84.33 & 62.49 & 25.89 \\
     Surrogate \cite{wu2018spatio} & 20& 64$\times$64  & 84.33 & 14.21 & 83.14 \\
     BNTT \cite{bntt} & 20& 32$\times$32   & 90.43 & 10.01 & 88.93 \\
       BNTT \cite{bntt} & 20& 64$\times$64   & 90.43 & 10.01 & 88.93 \\
        \bottomrule
        \\
    \end{tabular}%
    }
    \label{table:3232crossbar}
     \vspace{-4mm}
\end{table*}

\subsection{Accuracy comparison with crossbar non-idealities }

Table \ref{table:performance} shows the classification accuracies and performance ($\Delta$) of ANNs and SNNs with the impact of crossbar non-idealities.
From the experimental results, we observe the following:
(1) ANN-SNN conversion shows similar performance with the corresponding ANN model since both are basically identical in nature and functionality. The converted SNN has the same weights and its spike rates across all the time-steps are proportional to the corresponding float ANN activations.  
(2) On the other hand, non-idealities bring a huge performance drop for surrogate gradient learning based SNNs as compared to ANNs.
This is because surrogate gradient learning exploits spike dynamics (\ie spike timing, membrane potential, etc) which are highly susceptible to crossbar non-idealities. Note,
although surrogate gradient learning shows greater vulnerability on non-ideal crossbars, they can be implemented with a small number of time-steps.
(3) As the dataset and network architecture gets more complicated, the performance drop increases. For instance, ANN, Conversion, and Surrogate gradient learning show less than $1\%$ accuracy drop on MNIST. However, for CIFAR10 dataset, the performance degradation increases significantly. 
(4) We also examine the effect of Batch Normalization (BN) technique for SNN, called BNTT \cite{bntt}, which presents temporal BN parameters.
This technique enables SNNs to achieve high performance on software with even lesser number of time-steps.
Note, BNTT is based on surrogate gradient learning. 
Our empirical results show that BNTT aggravates the computation error arising from crossbar non-idealities.

\begin{figure}[t]
    \centering
    \includegraphics[width=0.9\linewidth]{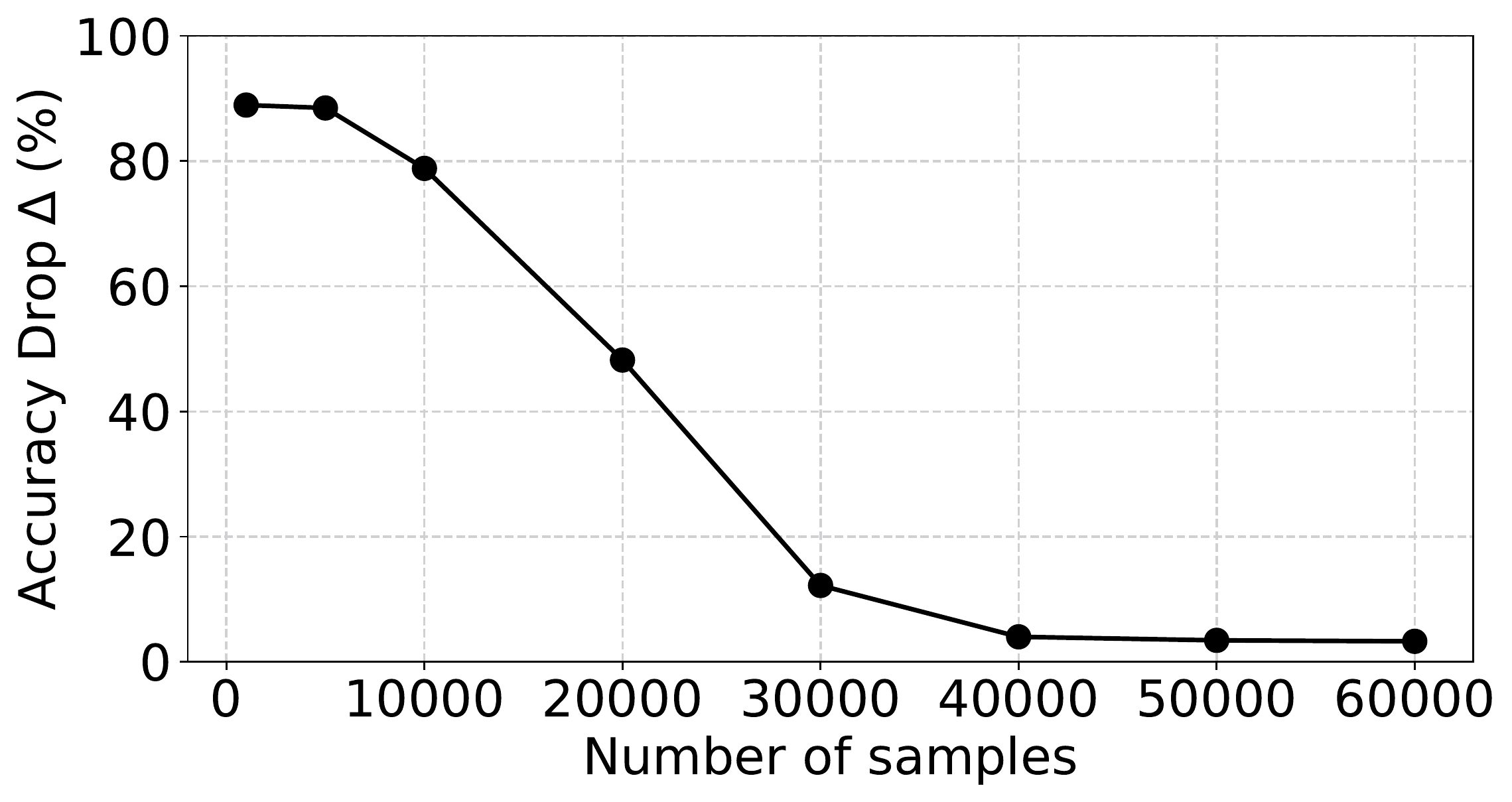}
    \caption{Effect of noise-aware BN adaptation on the performance of VGG9/CIFAR10 SNN, trained using BNTT, during inference on $64\times64$ crossbars.}
    \label{fig:bn_adpatation}
     \vspace{-3mm}
\end{figure}

\subsection{Ablation studies}

\noindent \textbf{Time-steps: }
We explore the performance drop with respect to time-steps for the various SNN learning algorithms on $64\times64$ non-ideal crossbars.
In Fig. \ref{fig:timesteps_abl}, we report the performance of surrogate gradient learning and ANN-SNN conversion with time-step \{10, 20, 30\} and \{100, 250, 500\}, respectively. 
For surrogate gradients, crossbar noise is accumulated across multiple time-steps, resulting in high performance drop at the high time-step regime.
For the ANN-SNN conversion case, since the converted SNN can be approximated with the ANN model, the performance remains consistent irrespective of the number of time-steps.

\noindent \textbf{Crossbar size: }
In addition to the previous experiments on  $64\times64$ crossbars, we evaluate the robustness of the SNN models on $32\times32$ crossbars, having lesser non-idealities (see Table \ref{table:3232crossbar}). 
We find that for BNTT, a huge performance drop occurs for even smaller crossbar size of $32\times32$, while SG learning attains superior performance as compared to $64\times64$ crossbars. This further validates the higher vulnerability of the BNTT-trained SNNs implemented on non-ideal crossbars. However, all performances are still inferior to the corresponding ANN models on comparing the $\Delta$ values for ANNs and SNNs in Table \ref{table:3232crossbar}). 






\subsection{Noise-aware BN adaptation for BNTT-trained SNNs}

From Table \ref{table:performance} and Table \ref{table:3232crossbar}, we find that BNTT shows a huge performance degradation on crossbars owing to the distribution mismatch between clean and noisy activations.
To address this, we present noise-aware BN adaptation which minimizes this mismatch by updating the average mean in the BN layers. 
Specifically, we forward a number of image samples through the SNN, adapting the moving average of BNTT with respect to noisy crossbar activations (while keeping other learnable parameters frozen). 
This can mitigate the impact of non-idealities on SNNs with BN techniques.
Fig. \ref{fig:bn_adpatation} shows the variation of performance drop ($\Delta$) with respect to the number of image samples forwarded during noise-aware BN adaptation.
The results show that SNN performance drop decreases with increase in the number of image samples, and with a large number of samples can surpass the corresponding ANN performance on crossbars.


\vspace{-2mm}

\section{Conclusion}

We explore the impact of crossbar non-idealites on the performance of SNNs. Interestingly, we find that SNNs trained using surrogate gradient learning are more vulnerable to crossbar non-idealites compared to ANNs due to repetitive crossbar computations across multiple time-steps. This also leads to the finding that SNNs evaluated with a small number of time-steps show lower performance degradation on non-ideal crossbars. Furthermore, applying batch-normalization technique on SNNs amplifies the effect of non-idealites on the activations, thereby resulting in greater performance losses. Unless such vulnerabilities pertaining to SNNs are addressed, the seamless integration of the SNN algorithms with emerging NVM-based hardware architectures will be precarious. Thus, in future we will explore SNN-crafted training algorithms using noise-aware training \cite{bhattacharjee2021neat} as well as managing the representation type/scheme of input spikes to crossbars \cite{kim2022gradient}, for mitigating the effect of non-idealites.

\vspace{-1mm}

\section*{Acknowledgement}
\vspace{-1mm}
This work was supported in part by C-BRIC, a JUMP center sponsored by DARPA and SRC, the National Science Foundation (Grant\#1947826), TII (Abu Dhabi) and the DARPA AI Exploration (AIE) program.

\vspace{-2mm}

{
\bibliographystyle{ieeetr}
\bibliography{egbib}
}

\end{document}